\documentclass{article}

\usepackage[final,nonatbib]{nips_2017}

\usepackage[utf8]{inputenc} 
\usepackage[T1]{fontenc}    
\usepackage{hyperref}       
\usepackage{url}            
\usepackage{booktabs}       
\usepackage{amsfonts}       
\usepackage{nicefrac}       
\usepackage{microtype}      
\usepackage{amsmath}
\usepackage{graphicx}
\usepackage{subcaption}

\title{Fusing Bird's Eye View LIDAR Point Cloud and Front View Camera Image for Deep Object Detection}

\author{
  Zining Wang\\
  Department of Mechanical Engineering\\
  University of California, Berkeley\\
  \texttt{wangzining@berkeley.edu}\\
  \AND 
  Wei Zhan\\
  Department of Mechanical Engineering\\
  University of California, Berkeley\\
  \texttt{wzhan@berkeley.edu}\\
  \AND 
  Masayoshi Tomizuka\\
  Department of Mechanical Engineering\\
  University of California, Berkeley\\
  \texttt{tomizuka@berkeley.edu}\\
}

\begin{document}

\maketitle

\begin{abstract}

We propose a new method for fusing LIDAR point cloud and camera-captured images in deep convolutional neural networks (CNN). The proposed method constructs a new layer called sparse non-homogeneous pooling layer to transform features between bird's eye view and front view. The sparse point cloud is used to construct the mapping between the two views. The pooling layer allows efficient fusion of the multi-view features at any stage of the network. This is favorable for 3D object detection using camera-LIDAR fusion for autonomous driving. A corresponding one-stage detector is designed and tested on the KITTI bird's eye view object detection dataset, which produces 3D bounding boxes from the bird's eye view map. The fusion method shows significant improvement on both speed and accuracy of the pedestrian detection over other fusion-based object detection networks. 

\end{abstract}

\section{Introduction}
LIDAR and camera are becoming standard sensors for self-driving cars and 3D object detection is an important part of perception in driving scenarios. 2D front view images from cameras provide rich texture descriptions of the surrounding, while depth is hard to obtain. On the other hand, 3D point cloud from LIDAR can provide accurate depth and reflection intensity information, but the resolution is comparatively low. Therefore, images and point cloud are complementary to accomplish accurate and robust perception. The fusion of these two sensors is a prerequisite for autonomous vehicles to deal with complicated driving scenarios. 

The recent progress of convolutional neural networks (CNN) for classification and segmentation has invoked particular interest in applying deep neural networks (DNN) for object detection. The DNN-based object detection with either LIDAR \cite{li2016vehicle,engelcke2017vote3deep} or camera \cite{chabot2017deep,cai2016unified,ren2017accurate} has been widely explored by researchers and pushed to a very high single-frame accuracy. However, 3D object detection is still hard for networks based on a single kind of sensor, which is shown in Table \ref{table1}. The camera-based network obtains high average precision on 2D image bounding box because of the rich texture information. Since a 2D camera does not contain accurate depth information, the 3D bounding box precision is very low for camera-based networks even with stereo vision \cite{chen2016monocular} and vehicle dimension prior knowledge \cite{chabot2017deep}. The state-of-the-art LIDAR-based networks have low precision in both 2D and 3D detections but the gap of precision between two detections is much lower compared to camera-based networks. It means that the objects detected by LIDAR are mostly accurate in terms of 3D localization. The reason for its low classification precision is the sparsity of the point cloud.

Sensor fusion gains the capability of achieving high accuracy in both 2D and 3D detections. Some fusion-based networks \cite{braun2016pose,DepthCNITSC2017} directly adapt from the Faster-RCNN structure with proposals from LIDAR so as to preserve the 3D information. Some \cite{kim2016robust,lange2016online} propose regions of interest from the front view but projects point cloud to the camera image plane for an early-stage fusion of information. The more recent MV3D \cite{chen2016multi} designs a complicated version of fusion after region proposal and gets the highest performance in 3D detection. All the networks to date do not fuse before 3D proposals of interested regions, which means that the fusion only helps the classification and regression stage. Fusion does not take place before highly potential objects are extracted by the region proposal.

In this paper, we propose a new camera and LIDAR fusion architecture in CNN that fuses different views before the region proposal stage. The fusion is allowed to happen before the proposal stage because it transforms the whole feature map instead of regions of interest(ROI), which is not presented in previous fusion-based networks. The main idea is to employ the point cloud and sparse matrix multiplication to transform feature maps in different views, such as bird's eye view and front view, efficiently. The method is integrated into an innovative sparse non-homogeneous pooling layer. A fusion-based detection network is constructed accordingly. The single-sensor data processing units are modularized and are quite flexible. Therefore, the CNN backbones can be directly adapted from the state-of-the-art networks developed for single sensor. In this paper, we are able to switch among VGGnet used by \cite{chen2016multi}, multi-scale network by \cite{cai2016unified} and 3D-CNN by \cite{Apple_VoxelNet}, and benefit from the best single-sensor network. 

Another contribution by the paper is that the new architecture fuses both LIDAR and camera information for 3D proposal, while other fusion-based networks merely use the LIDAR information. As a result, other networks have to fuse information after 3D regions proposals. When the fusion happens only after the region proposal part, the network is forced to use the old fashioned two-stage Faster-RCNN structure which is slow for real-time implementations. By fusing in an earlier stage, this paper gets rid of the "region proposal subnet" + "detection subnet" structure and benefits from the speed boost of recently developed one-stage detection networks like RetinaNet \cite{lin2017focal} and SSD \cite{liu2016ssd}.

\begin{table}[h]
  \centering
  \begin{tabular}{llllll}
    \toprule
   \multicolumn{2}{c}{Network} & \multicolumn{2}{c}{2D Detection} & \multicolumn{2}{c}{3D Detection}                \\
    \cmidrule{1-6}
    Type     & Name  &  Car & Pedestrian & Car & Pedestrian \\
    \midrule
    Stereo & Mono3D \cite{chen2016monocular} & 87.9 & 68.6 & 18.2 &-      \\
    Stereo & 3DOP \cite{mousavian20163d} & 88.6 & 67.5 & 9.5 &-      \\
    LIDAR & VeloFCN \cite{chen2016multi} & 70.7 & - & 32.1 & -     \\
    LIDAR & VxNet \cite{Apple_VoxelNet} & 86.0 & 44.1 & 65.1 & 33.7 \\
    Fusion & Pose-RCNN \cite{braun2016pose}    & 75.8 & 63.4 & - & -  \\
    Fusion & MV3D \cite{chen2016multi}    & 90.5 & - & 62.4 & -  \\
    
    \bottomrule
  \end{tabular}
  \caption{2D and 3D Detection Average Precision on KITTI dataset. Categories are of moderate difficulty.}
  \label{table1}
\end{table}

\section{Related Work}
\subsection{DNN for Object Detection} This section reviews the recent development of DNN-based detectors that are applicable to autonomous driving scenarios. One frame captured by sensors contains multiple objects involved in driving scenarios. As pioneered in the Selective Search methodology \cite{uijlings2013selective}, the network slides a window or cluster dense points on the feature map to produce some regions of interest (ROIs). It is called the region proposal stage, also referred to as the first stage in detection. The ROIs are then fed to a classifier to generate the classes and locations of objects, called the second stage. The first stage can be done by classic methods such as R-CNN and Fast-RCNN, or be done by another region proposal network in Faster-RCNN which improves the speed compared to previous ones.

Objects in the driving scenario have various sizes in the feature map due to distances and their inherent dimensions like cars, pedestrians and cyclists. Detecting from vastly different scales is hard for the vanilla Faster-RCNN. Feature pyramid using multi-scale features is implemented in MS-CNN \cite{cai2016unified}, FPN \cite{lin2016feature} and SubCNN \cite{xiang2017subcategory} as a solution.

The above two-stage network structures do not meet the time requirement in self-driving application. Recent works start to seek for one-stage detectors to avoid the overheads caused by the extra processing between two stages. SSD and YOLO \cite{redmon2016yolo9000} finish classification in the proposal stage and are much faster than two-stage detectors. As a trade-off, the accuracy trails due to class imbalance. The state-of-the-art RetinaNet addresses the problem with focal loss \cite{lin2017focal}.

\subsection{Detector Based on Single Sensor}
DNNs based on a single sensor have been developed in many other areas. Networks for image are very successful and their extensions to LIDAR are growing rapidly. Their CNN backbones serve as good candidates for the data processing units of a fusion-based network.

The camera-based detection receives the most attention from researchers. The competition is very intense on KITTI dataset\cite{geiger2012we} with hundreds of proposed structures. Most well-performed camera-based networks including MS-CNN, RRC and modified Faster-RCNN \cite{yang2016exploit} only produce high quality bounding boxes in 2D front view images. There are surprisingly some single-camera-based networks capable of 3D detection. By adding the prior  dimension knowledge of cars and cyclists, the earlier works such as 3DVP \cite{xiang2015data} and Mono3D are able to produce 3D bounding boxes. The more recent works such as \cite{chen2016monocular} Deep3DBox \cite{mousavian20163d} and DeepMANTA \cite{chabot2017deep} produce even more accurate 3D boxes from monocular vision than networks using stereo vision such as 3DOP \cite{chen20153d}.

The most intuitive thought of extending the successful 2D-CNN from images to the LIDAR point cloud is to use 3D-CNN with voxel representation. Unfortunately, the high computational complexity in the large outdoor scene makes it intractable. LIDAR data collected in autonomous driving scenario has its own inherent sparse property. Vote3Deep \cite{engelcke2017vote3deep} conducts sparse 3D convolution to reduce computational load, but it does not apply to GPUs with parallel acceleration. VoxelNet\cite{Apple_VoxelNet} implements 3D convolution on voxel representation that is coarsely divided in height. It achieves highest scores in 3D detection with acceptable speed. VeloFCN \cite{li2016vehicle} and SqueezeSeg \cite{wu2017squeezeseg} projects point cloud to the front view with cells gridded by LIDAR rotation and then applies normal 2D CNN for classification and segmentation. The front view representation of point cloud shares the same multi-scale problem as camera, because the sizes of objects change as distance varies.

\subsection{Fusion-Based Detector}
The network based on camera and LIDAR fusion, especially for pedestrian detection, has not been sufficiently investigated. Some works \cite{kim2016robust,lange2016online} apply early fusion by projecting point cloud to the image plane and augment the image channels after upsampling. Such structure fuses camera and LIDAR data through the whole network but only on image plane. It means that the accurate 3D information of LIDAR point cloud is almost lost. The localization banks on that the regression can magically retrieve the 3D measurement of point cloud, which is not the case according to their low 3D detection score. Pose-RCNN \cite{braun2016pose} adapts the Fast-RCNN structure where the region proposal is done by classic selective search in LIDAR voxel representation. The fusion structure does not feed LIDAR data into the deep convolutional network which means the classification relies merely on camera data. The state-of-the-art MV3D network applies region proposal network (RPN) on the point cloud projected to the bird's eye view plane. It preserves the 3D measurement in the region proposal stage. Then it uses the Faster-RCNN and Deep-fused Net \cite{wang2016deeply} structure in the second stage. Note that the region proposal stage only takes the bird's eye view LIDAR data. The quality of proposals is limited by using single view and single sensor shown in Section \ref{QualityofProposals}. The speed is also limited because the fusion must happen at the second stage. In this paper, we propose an efficient fusion scheme working in the first stage. The proposal takes into account the information from both CNN-processed bird's eye view LIDAR and front view camera data. This structure not only produces high-quality proposals by camera and LIDAR fusion but also allows building fast one-stage fusion-based detectors.

\section{Sparse Non-homogeneous Pooling with LIDAR Point Cloud}
\label{Sparse_pooling}
The sparse non-homogeneous pooling is a method that transforms two feature maps $f( {x,y} ),g( {u,v} )$,  where the transformation is linear but non-homogeneous. For example, the general convolution and pooling used in CNNs are homogeneous since \[f\left( {x,y} \right) = \sum\limits_{u,v} {k\left( {x-u,y-v} \right)g\left( {u,v} \right)} \] The kernel $k(u,v)$ has the same finite support on $(u,v)$, such as $[-1,1]\times[-1,1]$ with the kernel size of 3, which is independent of $(x,y)$. The spatial transformer network \cite{jaderberg2015spatial} is also homogeneous as it predicts a uniform transformation matrix for the whole feature map. In general, the front view map and bird's eye view map are related by the projective transformation 
\[\left[ {\begin{array}{*{20}{c}}
{uw}\\
{vw}\\
w
\end{array}} \right] = P\left[ {\begin{array}{*{20}{c}}
x\\
y\\
z\\
1
\end{array}} \right]\]

where $P$ is the projection matrix derived from camera-LIDAR calibration, $(u,v)$ is the pixel in image and $(x,y)$ is the coordinate in bird's eye view map. $z$ is the additional height coordinate. If we want to transform a feature map from image to bird's eye view, the transformation is
\[f\left( {x,y} \right) = \sum\limits_{u,v} {{k_{x,y}}\left( {u,v} \right)g\left( {u,v} \right)} \]
and the support of kernel is $\sup \left( {{k_{x,y}}} \right) = \left\{ {\left( {u,v} \right)\left| {{{\left[ {u{\kern 1pt} {\kern 1pt} ,v} \right]}^T} = {w^{ - 1}}{P_{12}}X,w \in {\mathbb{R}^ + }} \right.} \right\}$ where $X=[x,y,z,1]^T$ and $P_{12}$ contains the first two rows of $P$. Both the elements and measure of the support depend on $(x,y)$. The support is usually a line. If the transformation was done as above, the computation would be heavy and non-parallel because a large but various number of $g(u,v)'s$  was involved for each $(x,y)$. The proposed sparse non-homogeneous pooling uses the point cloud to reduce the support region. It also formulates the transformation as sparse matrix multiplication so that pooling of the full future map is done in one matrix multiplication.
 

\subsection{Bird's Eye View LIDAR representation}

The most important coordinate information of 3D object detection in autonomous driving scenarios is the x, y and orientation on the ground. In fact, the height of objects can be easily estimated from the ground. Bird's eye view takes the ground plane for the construction of feature map and thus provides the best proposal quality on x,y axis which are more difficult to estimate than z axis. Currently, no front view detector achieves good performance in terms of 3D bounding box. In bird's eye view representation, point cloud is discretized into a $L_b\times{W_b}\times{H_b}$ grid. The $L_b\times{W_b}$ grid on the ground is dense and $H_b$ is small so that the number of cells is similar to a 2D grid instead of 3D dense voxel representation. 

There are two popular kinds of input features encoded in each cell. One is the hand-crafted feature such as density, height, reflection and position used in MV3D. The dimension is about 10 for each cell. Choosing what feature to use can be a difficult hyper-parameter tuning task. The other one is the network created feature by processing all points inside the cell, such as the PointNet structure\cite{Garcia2016PointNet} called voxel feature encoding (VFE) layer in VoxelNet. Each cell possesses a 128 dimensional feature learned by the network. In this paper, as we switch from different LIDAR data processing units, the input features are changed accordingly. As long as the features are arranged as gridded cells, the sparse non-homogeneous pooling method remains unchanged.


\subsection{Non-homogeneous Pooling as Sparse Matrix Multiplication}
\label{Non-homo}
The transformation between front view (image) map and bird's eye view map can be sparsified by the point cloud. Instead of matching one pixel $(u,v)$ in front view with a full homogeneous line $(\lambda{x},\lambda{y})$ in bird's eye view, only $(u,v)$ and $(x,y)$ that share the same point in the point cloud are paired. The transformation is still non-homogeneous as the pairing does not guarantee one-to-n (n fixed) mapping, but the computation is sparse as the number of points in point cloud is of only 10,000 scale. Suppose the front view map is of size $H_f\times{W_f}$ and the bird's eye view map is $L_b\times{W_b}$. The LIDAR point cloud is $\left\{ {\left( {{x_i},{y_i},{z_i}} \right)\left| {i = 1,2, \cdots N} \right.} \right\}$, the transformation kernel is sparsified as
\[\begin{gathered}
  {k_{x,y}}\left( {u,v} \right) = {\delta _{\left( {x,y} \right)\left( {{x_i},{y_i}} \right)}}{k_{x,y}}\left( {u,v} \right){\delta _{\left( {u,v} \right)\left( {{u_j},{v_j}} \right)}}, \hfill \\
  {\kern 130pt}  i,j = 1,2 \cdots N \hfill \\ 
\end{gathered} \]
because only $(u,v), (x,y)$ correspond to a LIDAR point are transformed. The transformation becomes
\[\begin{aligned}
  f\left( {{x_i},{y_i}} \right) &= \sum\limits_{u,v} {{k_{{x_i},{y_i}}}\left( {u,v} \right){\delta _{\left( {u,v} \right)\left( {{u_j},{v_j}} \right)}}g\left( {u,v} \right)}  \\ 
   &= \sum\limits_j {{k_{{x_i},{y_i}}}\left( {{u_j},{v_j}} \right)g\left( {{u_j},{v_j}} \right)}  \\ 
   &= \sum\limits_j {{k_{{x_i},{y_i}}}\left( {{u_j},{v_j}} \right){\delta _{\left( {{x_i},{y_i}} \right)\left( {{x_j},{y_j}} \right)}}g\left( {{u_j},{v_j}} \right)}  \\ 
\end{aligned} \]
where 
\[{\delta _{ab}} = \left\{ {\begin{array}{*{20}{l}}
  {1,a \sim b} \\ 
  {0,{\text{otherwise}}} 
\end{array}} \right.\]
So the transformation kernel, instead of having a small support like convolution, is a sparse ${L_bW_b}\times{H_fW_f}$ matrix of $\left\{ {\left( {x,y} \right)} \right\} \times \left\{ {\left( {u,v} \right)} \right\}$, or of ${{{\mathbb{Z}_N}} \mathord{\left/{\vphantom {{{\mathbb{Z}_N}}  \sim }} \right.\kern-\nulldelimiterspace}  \sim } \times {\mathbb{Z}_N}$ where $ \sim $ is the equality and $[\cdot]$ means round to integer. \[\left( {{x_i},{y_i}} \right) \sim \left( {{x_j},{y_j}} \right) \Leftrightarrow \left( {{[x_i]},{[y_i]}} \right) = \left( {{[x_j]},{[y_j]}} \right)\] 

There are at most $N$ non-zero elements in the sparse matrix regardless of the size of feature maps. The kernel is normalized by the number of points in one cell. 
\[k_{x_i,y_i}(u_j,v_j) = \left| {\left\{ {k \in \left\{ {1,2, \cdots N} \right\}\left| {\left( {{x_i},{y_i}} \right) \sim \left( {{x_k},{y_k}} \right)} \right.} \right\}} \right|^{-1}\]
Note this can be extended to more general interpolation methods like bilinear pooling by associating the LIDAR point with not only the one pixel it projects to, but also its neighbors and normalize the matrix row-wise to have sum 1.

\begin{figure}[h]
  \centering
  \includegraphics[width=0.8\textwidth]{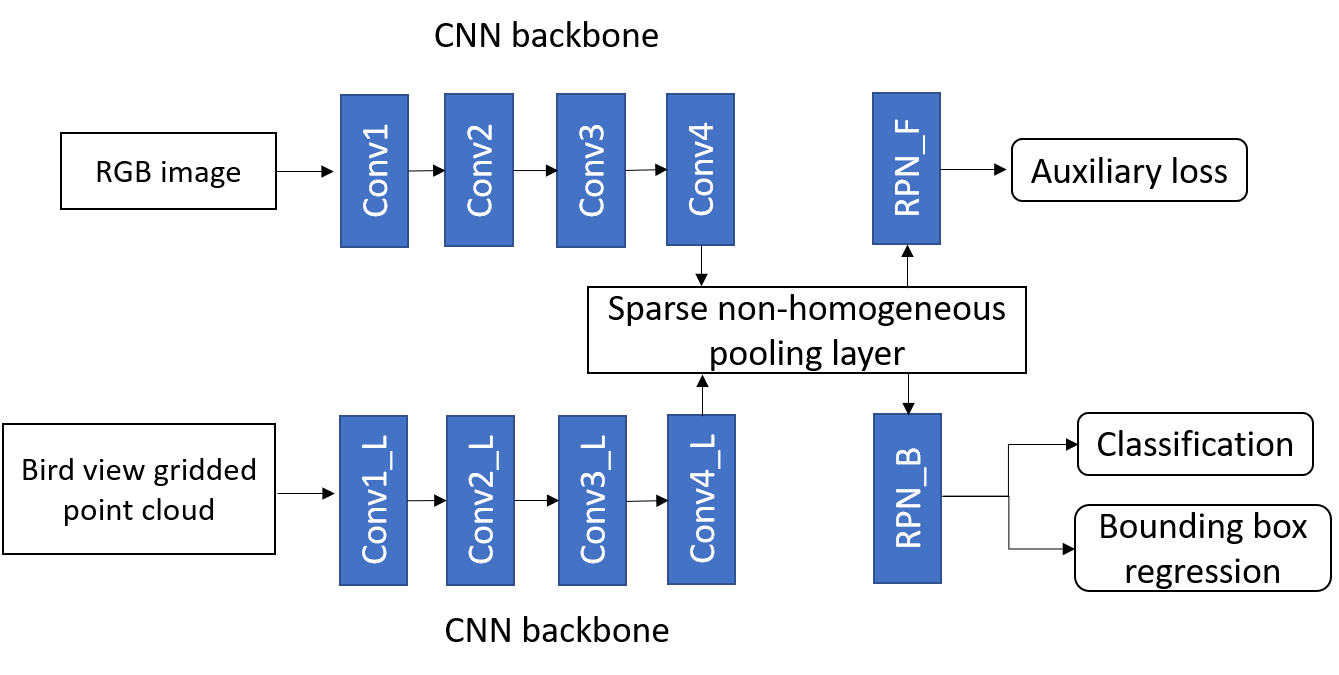}
  \caption{The vanilla fusion-based one-stage object detection network.}
  \label{fig_MV3D_img}
\end{figure}

\subsection{Fusion of Front View and Bird's eye View Features}

\begin{figure}[h]
  \centering
  \includegraphics[width=0.8\textwidth]{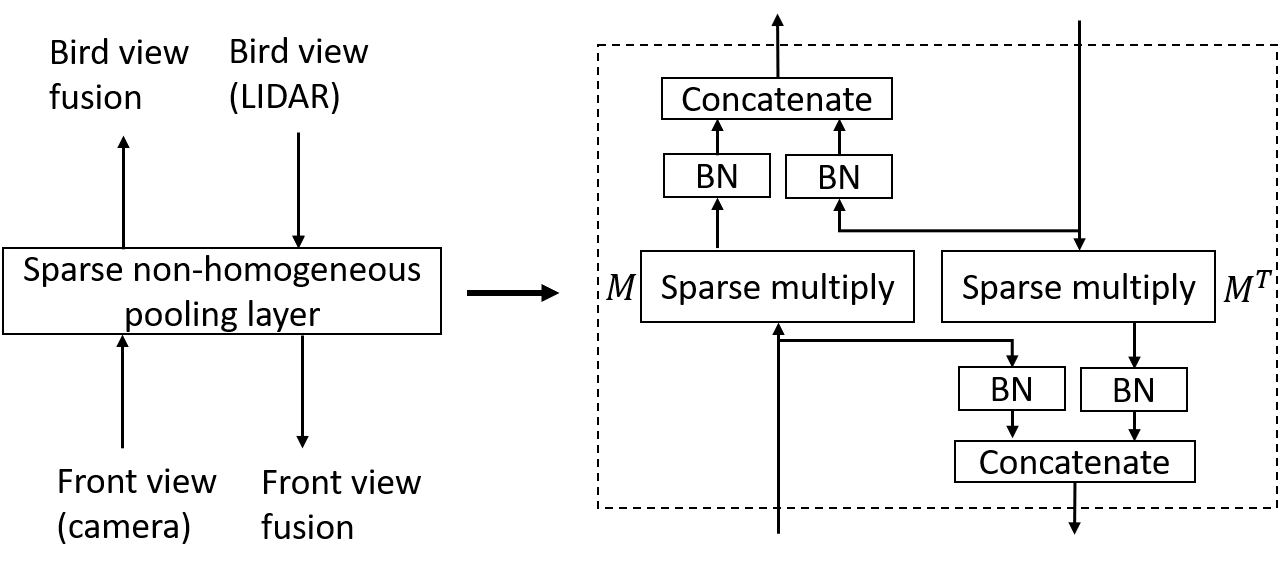}
  \caption{The sparse non-homogeneous pooling layer that fuses front view image and bird's eye view LIDAR feature.}
  \label{SHPL}
\end{figure}

Section \ref{Non-homo} introduces the case of transformation from front view $(u,v)$ to bird's eye view $(x,y)$ but this can also be done from bird's eye view to front view by just exchanging the coordinates. When applying the non-homogeneous pooling in network, since all the coordinate of the feature maps and the calibration matrix $P$ are known, the pairs and matrix can be pre-calculated for each frame. There is no parameter to train for the pooling operation. The structure of the layer is illustrated in Figure \ref{SHPL}

The sparse non-homogeneous pooling layer takes the feature map, such as the image, as input and $(u_i,v_i), (x_i,y_i)$ pairs as parameters. The layer constructs a ${L_bW_b}\times{H_fW_f}$ sparse matrix $M$. The $H_f\times{W_f}\times{C}$ feature map is flatten to a dense $H_f{W_f}\times{C}$ matrix $F$. Then the pooled feature map is $B=MF$ of ${L_bW_b}\times{C}$ and can be concatenated with the target feature map for further processing. Batch normalization (BN) is applied for both feature maps before concatenation.

\begin{figure*}[h!]
    \centering
    \begin{subfigure}[h]{0.8\textwidth}
        \centering
        \includegraphics[height=1.4in]{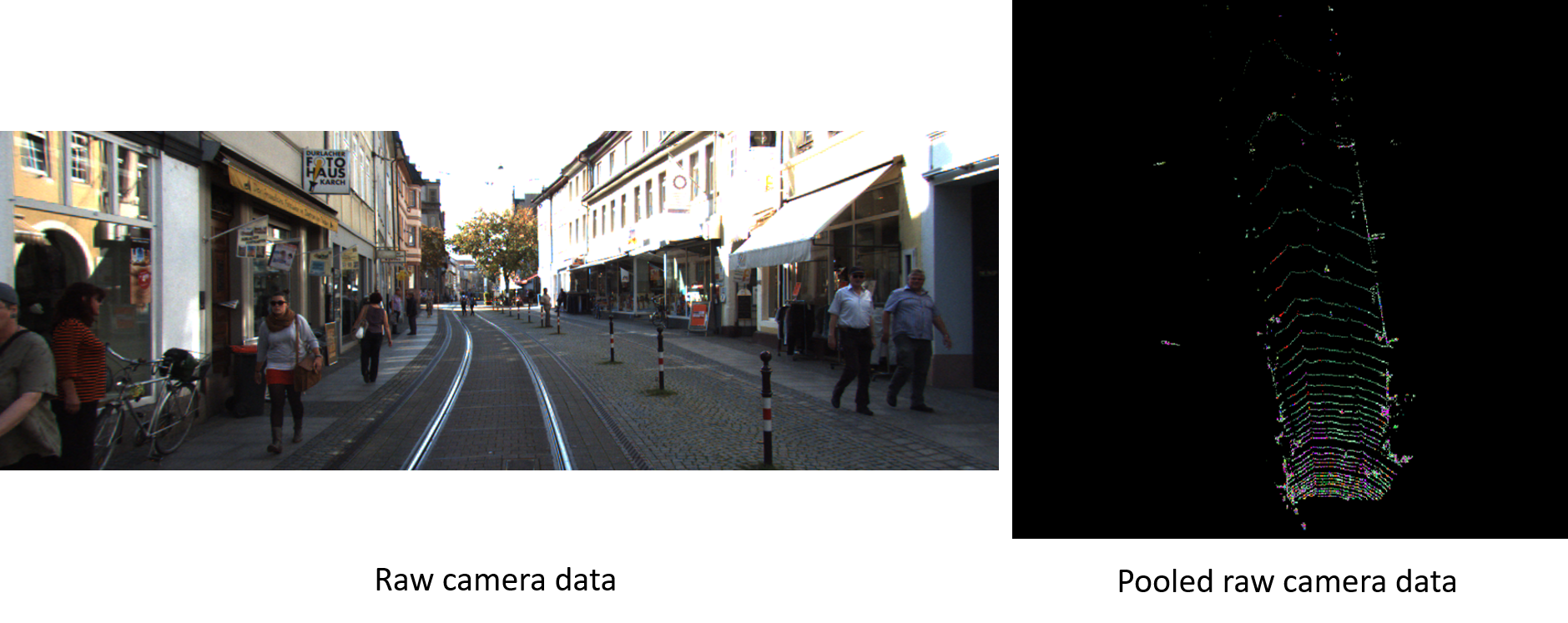}
        \caption{ }
    \end{subfigure}
    ~ 
    \begin{subfigure}[h]{0.8\textwidth}
        \centering
        \includegraphics[height=1.4in]{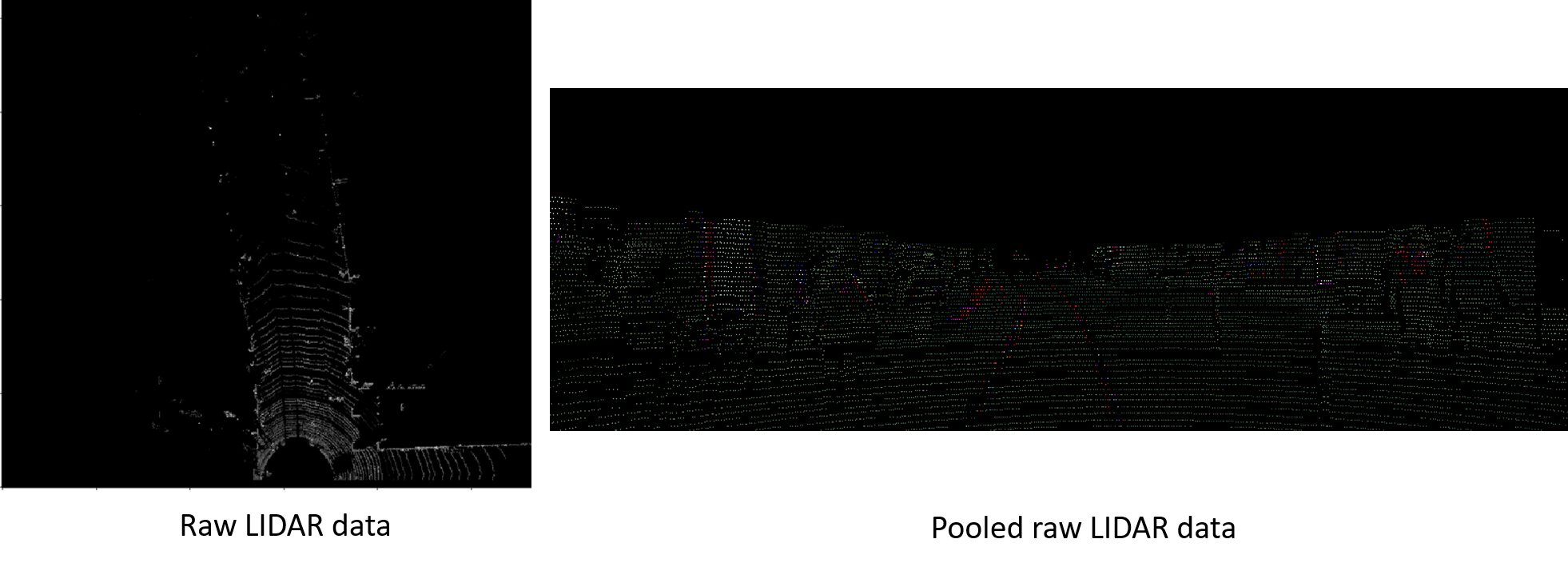}
        \caption{ }
    \end{subfigure}
    ~ 
    \begin{subfigure}[h]{0.8\textwidth}
        \centering
        \includegraphics[height=1.4in]{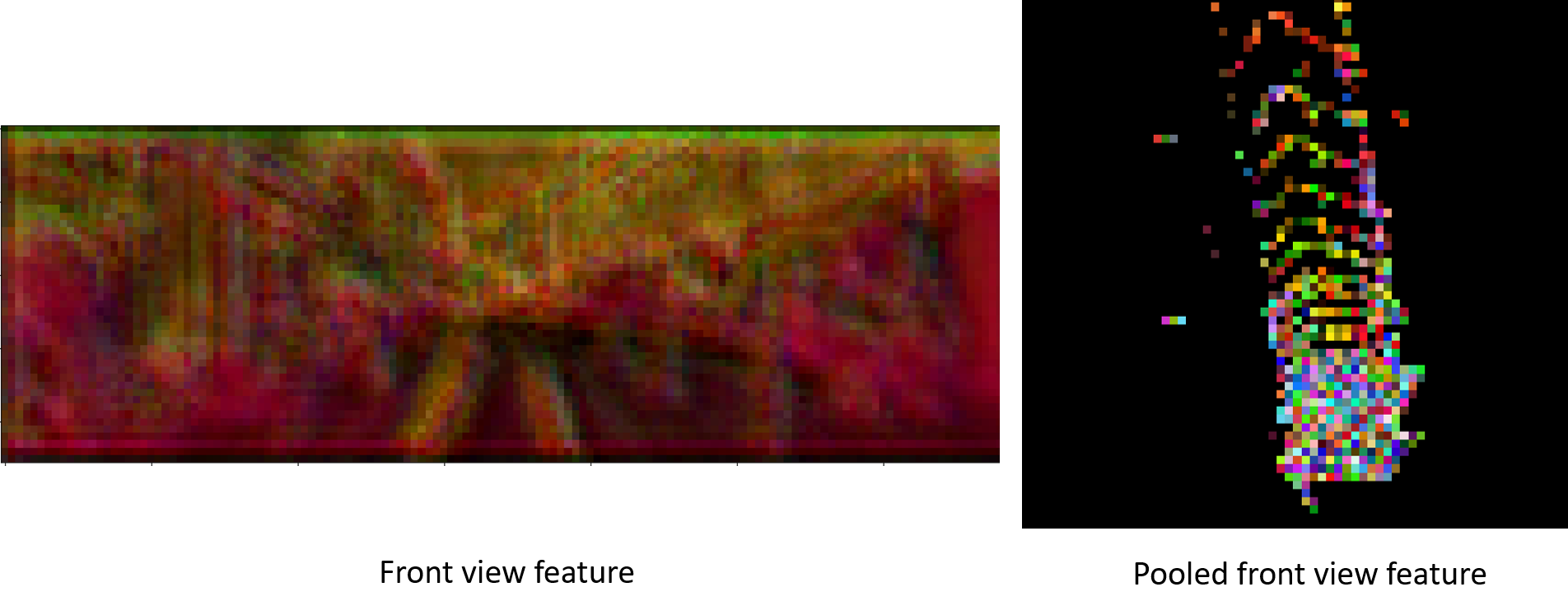}
        \caption{ }
    \end{subfigure}
    ~ 
    \begin{subfigure}[h]{0.8\textwidth}
        \centering
        \includegraphics[height=1.4in]{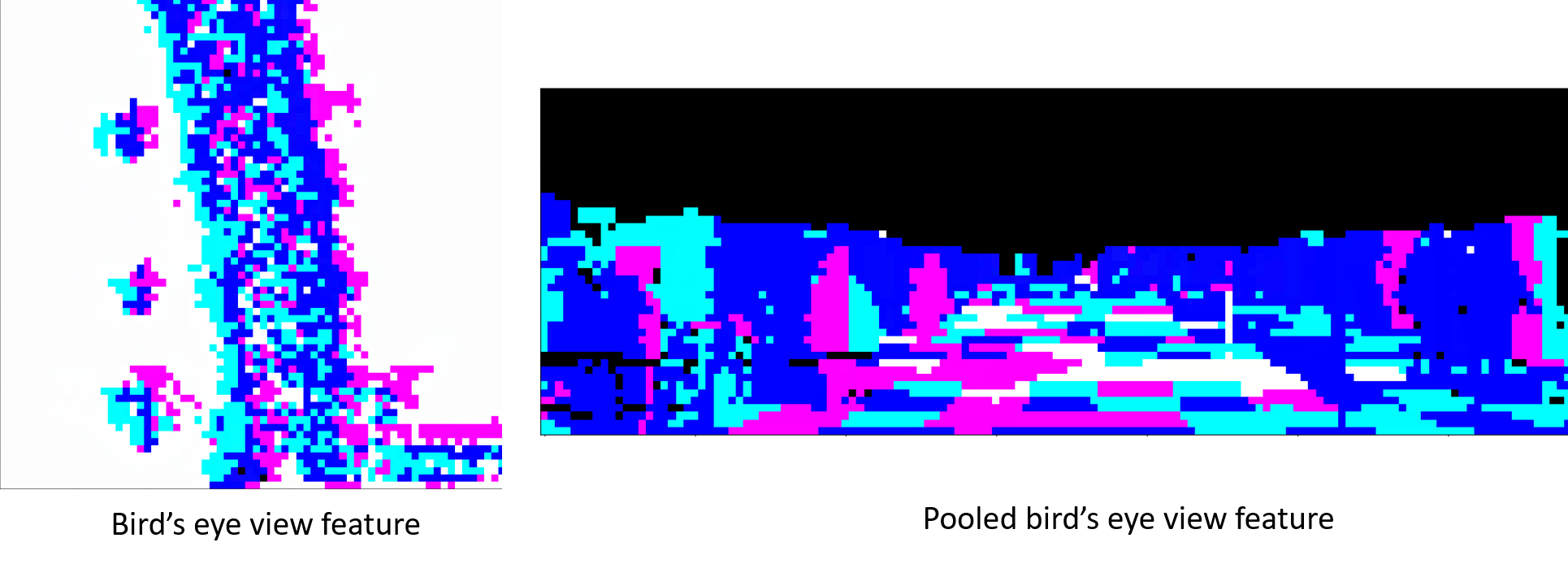}
        \caption{ }
    \end{subfigure}
    \caption{The sparse non-homogeneous pooling example: (a)From camera image to bird's eye view. (b)From bird's eye view to camera image. (c)From front view conv4 layer to bird's eye view conv4 layer. (d)From bird's eye view conv4 to bird view conv4.}
    \label{fig_pool}
\end{figure*}

The fusion between bird's eye view and front view feature maps is not recommended in the early stage. Due to the low resolution of point cloud compared to the camera image, pooling of raw input results in a very low usage of image data shown in Figure \ref{fig_pool}. 20,000 points in the point cloud only results in $0.4\%$ usage of pixels in raw RGB image in KITTI dataset. The sparse pooling is preferred to be applied to deeper layers like the network shown in Figure \ref{fig_MV3D_img}. When the pooling is done at the conv4 layer where the raw input is downsampled by 8, most of the front view and bird's eye view features are involved in fusion except for the part of sky with no laser reflection.

\section{One-Stage Fusion-Based Detection Network}
\label{One_stage}
The fusion structure introduced in Section \ref{Sparse_pooling} allows one-stage detector because the fusion is done in the first stage across different views, unlike \cite{chen2016multi} where fusion is done only after RoI pooling in the second stage. In this section we introduce a one-stage detector that takes front view camera image and bird's eye view LIDAR point cloud as input and produces 3D bounding box from bird's eye view without RoI pooling. The detection scheme is adapted from \cite{lin2017focal} but those of \cite{liu2016ssd,redmon2016yolo9000} are also compatible.

\subsection{3D Region Proposal with Fusion structure}

The network structure is shown in Figure \ref{fig_MV3D_img}. There are two fully convolutional backbones, namely the image unit and LIDAR unit. The sparse non-homogeneous pooling layer serves as the cross-bridge of two units to exchange information between sensors. The image unit uses the same RPN structure as recent camera-based one-stage detectors. Although the region proposal is not used during test, an auxiliary loss is still applied on the image CNN so that image features get supervision from the label in front view in addition to the that from the 3D proposal. It serves the similar functionality as the auxiliary loss in \cite{chen2016multi}.

Since the region proposal is in the bird's eye view, objects of different distances are of the similar size if they are in the same category. The vehicles in KITTI dataset have bounding box size $(l,w,h)$ of (4.0, 1.6, 1.6)m $\pm$ (0.6, 0.1, 0.2)m and pedestrians are of (0.9, 0.6, 1.6)m $\pm$ (0.2, 0.1, 0.2)m in the bird's eye view plane. There is no multi-scale issue hence the multi-scale structure in \cite{chen2016multi,lin2016feature} are not needed. 

\begin{figure}[h]
  \centering
  \includegraphics[width=0.8\textwidth]{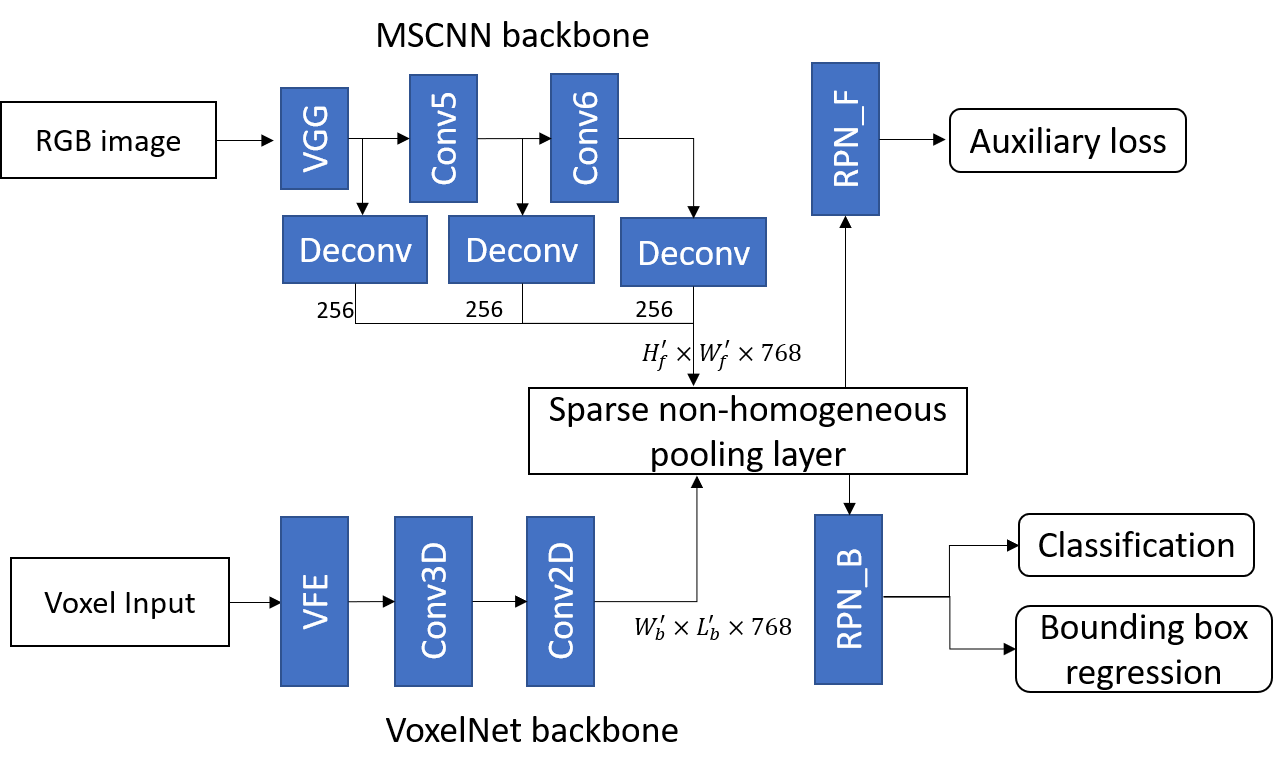}
  \caption{The fusion-based one-stage object detection network with state-of-the-art single-sensor networks.}
  \label{fig_Voxel_img}
\end{figure}

\subsection{One-Stage Object Detection}
The resolution of image and bird's eye view point cloud captured in the autonomous driving scenario is much larger than the benchmark datasets used to evaluate general detection networks. The test time per frame is important for practical application. One-stage detectors are much faster than two-stage detectors because they do not have the RoI pooling, non-minimum suppression and fully connected operation in the second stage. One-stage detectors directly predict bounding boxes from all the proposals produced by RPN whose number is usually of 100K scale, which means the proposals should be of very high quality to extract the very few true positives. 

\subsubsection{Class Imbalance and Hard Negative Mining}

The class imbalance problem in detectors based on selective search \cite{uijlings2013selective} and its solution in DNNs is summarized in \cite{lin2017focal}. For two-stage detectors \cite{ren2015faster}, the imbalance problem is addressed by RoI pooling as it keeps a positive:negative$\approx$1:3 ratio in the second stage. Actually, other second-stage detectors like \cite{chen2016multi} also use bootstrapping and weighted class loss in the first stage when the class imbalance is huge for small objects. 

Class imbalance problem is because in addition to the prediction of probability, classification also involves binary (or discrete) decision \cite{Frank_statistical_thinking}. A utility function or decision cost is imposed to the classification result where many true negatives, whose probabilities are below some threshold, contribute almost no cost. The so called hard negative mining solutions share the same idea of lifting the loss of hard negatives, because there is a mismatch between the used cross-entropy loss and the decision loss that actually measures the performance. This work uses the focal loss \cite{lin2017focal} which is a weighted sum of cross-entropy \[{\text{FL}}\left( {{\mathbf{p}},y} \right) = {\alpha _y}{\left( {1 - {p_y}} \right)^\gamma }{\text{CE}}\left( {{\mathbf{p}},y} \right)\]
${\mathbf{p}}$ is the vector of probability of the sample among all classes. $y$ is the label of the sample. $p_y=\mathbf{p}(y)$ is the probability at class $y$ and $\alpha{_y}$ is an empirical weight for each class. ${\text{CE}}( \cdot )$ is the cross-entropy. Focal loss is very efficient and is demonstrated to be effective on image-based one-stage object detector by \cite{lin2017focal}.

\subsubsection{Pedestrian Labeling in Bird's Eye View Map} 
\label{ped_label}

MV3D is very successful on the 3D detection of vehicles by outperforming all other methods on KITTI by a large margin. However, it does not have detection score on the pedestrian and cyclist class. On the one hand, the reason is that vehicles have distinguishable features like square corners, but pedestrians are only dense dots similar to trees and traffic signs on the bird's eye view map. On the other hand, the sizes of pedestrians on the bird's eye view map are very small. The general size of a pedestrian is $1m\times0.6m$ which is only $10\times6$ with a resolution of $0.1m$ on a $600\times600$ bird's eye view gridded map. Even with a down-sample rate of 4, its anchor becomes $2.5\times1.5$. One shift will cause the IoU (interaction over union) off 0.5. If rotation is taken into account, the IoU will fall even lower. To address the labeling of small objects like pedestrians in bird's eye view map, anchor boxes and ground truth boxes are rotated to axis-aligned boxes for IoU calculation. This means the IoU only considers the size difference between the anchor and ground truth. Then the angle and position differences are calculated as regression targets.

\section{Experiments}
\label{Experiment}

The network based on sparse non-homogeneous pooling is evaluated on the KITTI dataset which has calibrated camera and LIDAR data and ground truth 3D bounding boxes. The KITTI 3D object and bird's eye view evaluation are used. We focus on the pedestrian category as the maximum average precision (mAP) on KITTI is still very low among all fusion-based network where average precision (AP) is only about 26\%. It is shown that the fusion structure helps a lot on the 3D proposal of pedestrians from bird's eye view since there is enough fused information. 

\subsection{Implementation}
\subsubsection{Network Details}
Two kinds of CNN backbones are used for the network. The first one uses VGG for both LIDAR and camera, called the vanilla version shown in \ref{fig_MV3D_img}. The VGG16 is full VGG pretrained on ImageNet in contrast to the reduced VGG used in \cite{chen2016multi}. The input is the $1280\times{384}$ camera image and the $600\times{600}\times{9}$ gridded LIDAR bird's eye view representation with 0.1m resolution on the ground. The feature map produced by the backbone is down-sampled by 4 times in bird's eye view and 8 times in front view. The second one uses MS-CNN\cite{cai2016unified} for camera and VoxelNet\cite{Apple_VoxelNet} for LIDAR. The input is the $1280\times{384}$ camera image and LIDAR voxels following the same setting as VoxelNet. The feature map produced by the backbone is down-sampled by 2 times in bird's eye view and 8 times in front view. For both kinds, camera inputs are augmented by globally scaling randomly between $[0.95,1.05]$ and shifting randomly between $[-10,10]$. LIDAR inputs are augmented following the instruction of VoxelNet. The calibration matrices for sparse pooling is changed accordingly. The network is implemented using TensorFlow and one TITAN X GPU as hardware.

The focal loss is applied to all anchors to deal with the class imbalance. Different from \cite{lin2017focal}, it is observed that using the focal loss for the non-object class only and cross-entropy for object classes performs better than using focal loss for all classes. Moreover, instead of using biased initial weights to prevent the instability of training in \cite{lin2017focal}, an adaptive weight of losses is used. For non-object class, the loss is formulated as 
\[loss_{neg} = (1-\alpha)\times \text{CE}_{neg}+\alpha\times \text{FL}_{neg}\]
$\text{CE}_{neg}$ is the common cross entropy loss and $\text{FL}_{neg}$ is the focal loss on all non-object anchors . $\alpha = 0$ for the first 10\% of iterations and $\alpha = \text{recall}_{pos}$ which is the average recall rate of objects updated every 500 iterations by exponential average with a decay rate of 0.998. As the recall rate grows higher during training, the weight of focal loss becomes higher, bringing higher weight on hard negatives. This is similar to that used in \cite{cai2016unified} where random sampling is used on all negative anchors first and bootstrapping is used later. The same smooth $l_1$ loss as \cite{girshick2015fast} is used for bounding box regression.

\subsubsection{Evaluation Speed}
The sparse non-homogeneous pooling layer is tested to be efficient. The pooling of raw inputs (camera image and bird's eye view LIDAR above) takes 20ms while the pooling of features produced by VGG16 ($155\times{46}\times{512}$ for camera and $75\times{75}\times{512}$ for LIDAR) takes 14ms. With the efficient sparse non-homogeneous pooling fusion and one-stage detection structure \ref{fig_MV3D_img}, the test time is 0.11s per frame compared to the 0.7s per frame in MV3D. With VoxelNet, the inference time is longer because there is no available official version of VoxelNet. The implementation is modified from an unofficial publicly available reproduction by Jeasine Ma and does not reach the speed claimed by the paper \cite{Apple_VoxelNet}. The test time consists of the sparse matrix construction, network inference and bounding box post-processing (NMS) but the file I/O is excluded.  

The network is evaluated on the KITTI object detection benchmark using the same train/validation split as provided in \cite{cai2016unified}. The bird's eye view average precision is used for evaluation. Actually the AP on bird's eye view benchmark is close to that on the 3D benchmark of KITTI for all methods involved.

\subsection{Quality of Proposals}
\label{QualityofProposals}
To verify the fusion structure is effective augmenting the feature in the RPN, the recall and precision in the RPN with VGG16 is compared with that of the MV3D structure. To make fair comparison, the cross-entropy loss is used in RPN instead of focal loss and IoU threshold is set to 0.5. The MV3D implemented in this paper is unofficial but reproduced according to the paper. Table \ref{table2} shows, for vehicles, the precisions are similar but for pedestrians, the fusion-based RPN has much higher precision. This is because the pedestrian is hard to distinguish with just LIDAR data, while vehicles are distinct from bird's eye view. The proposals for pedestrians require more information and the fusion structure provides the front view image feature, making proposal quality of pedestrians similar to that of vehicles.
\begin{table}[h]
  \centering
  \begin{tabular}{lllll}
    \toprule
   Network & \multicolumn{2}{c}{Vehicle} & \multicolumn{2}{c}{Pedestrian}                \\
    \cmidrule{1-5}
    Type     &  Recall & Precision & Recall & Precision \\
    \midrule
    MV3D & 99.4 & 17.3 & 97.8 & 4.2   \\
    ours     & 99.2 & 19.5 & 96.6 & 17.3      \\
    \bottomrule

  \end{tabular}
  \caption{RPN performance of RPN on pedestrian and vehicles on the validation set. All difficulty levels are included}
\label{table2}
\end{table}

\subsection{Detection Results and Discussion}

\begin{figure*}[h]
  \centering
  \begin{subfigure}[h]{0.5\textwidth}
        \centering
        \includegraphics[height=1.8in]{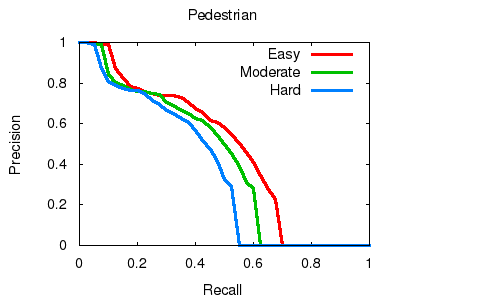}
        \caption{}
    \end{subfigure}
    ~ 
    \begin{subfigure}[h]{0.4\textwidth}
        \centering
        \includegraphics[height=1.8in]{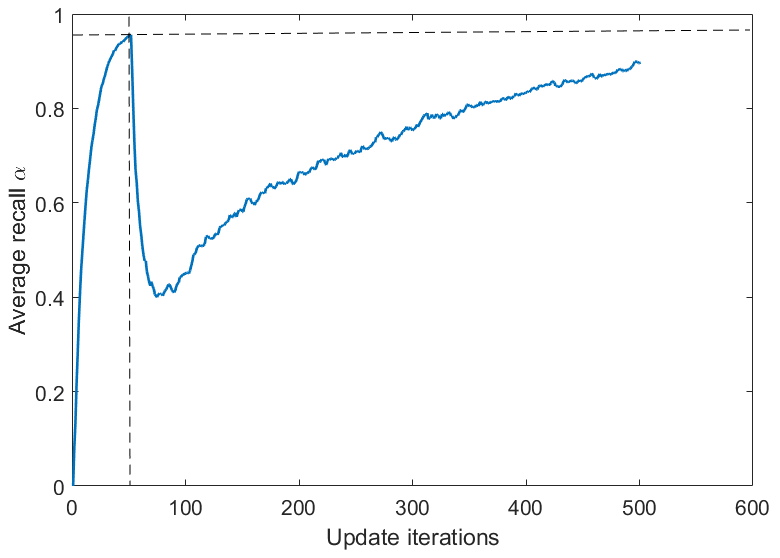}
        \caption{ }
    \end{subfigure}
  \caption{(a) The PR-curve of our fusion-based network on the validation set. (b) The weight of focal loss within 30 epochs}
  \label{result_PR1}
\end{figure*}

The pedestrian detection result is evaluated on the validation dataset and shown in Figure \ref{result_PR1}. Due to the limit of hardware, the network is not trained to the same number of epochs as in MV3D and VoxelNet which are around 150 epochs. The network is trained for 30 epochs on the train split with 3712 samples and evaluated on the validation split with 3769 samples. The learning rate is 0.0005 for the first 50\% iterations and reduces linearly to 0 for the last 50\%. The CNN backbone of LIDAR part is initialized randomly and the camera part loads the pretrained weights from VGG and MS-CNN, for vanilla version and VoxelNet+MS-CNN version respectively.

Table \ref{table3} shows the performance comparison of 3D detection networks presented on the KITTI dataset. The VoxelNet, due to its superior LIDAR input representation, has much higher score than other networks. It is even better than our vanilla version fusion-based network because the vanilla version uses the same hand-crafted LIDAR input as MV3D. By fusing VoxelNet and MS-CNN with the proposed sparse non-homogeneous pooling layer and one-stage detection network, we are able to achieve the highest performance on the validation set. Figure \ref{result_PR1}(b) shows the curve of average recall rate during training. The recall drops when the focal loss is activated and is still increasing at the end of the training process.

\begin{table}[h]
  \centering
  \begin{tabular}{llll}
    \toprule
   Network & \multicolumn{3}{c}{Pedestrian}              \\
    \cmidrule{1-4}
    Name     &  Easy & Moderate & Hard  \\
    \midrule
    3dssd & 27.4 & 24.0 & 22.4   \\
    AVOD & 34.4 & 26.1 & 24.2 \\
    VoxelNet(Official) & 46.1 & 40.7 & 38.1 \\
    Vanilla     & 34.0 & 31.4 & 29.3       \\
    VoxelNet(Reproduced) & 46.9 & 38.1 & 33.9 \\
    VoxelNet+MS-CNN   & \bf 51.3 & \bf 45.0 & \bf 40.2       \\
    \bottomrule

  \end{tabular}
  \caption{Comparison with the current state of the art on KITTI. Note the last three rows are on the validation set with 30 epochs and the first three rows are on the test set with 150 epochs.}
\label{table3}
\end{table}

\section{Conclusion}

We proposed a sparse non-homogeneous pooling method to efficiently transform and fuse features from different views of LIDAR point cloud and images from cameras. The proposed method enabled the fusion before the proposal stage so that the whole feature map can be exploited and fused, which was not presented in previous fusion-based networks. High-quality 3D proposals were produced by the corresponding one-stage detector designed. The detector achieved the best performance in the 3D detections of pedestrians from bird’s eye view on KITTI dataset, which was significantly superior to the state-of-the-art fusion-based detectors.

\bibliographystyle{unsrt}
\bibliography{reference}

\end{document}